\documentclass[sn-mathphys]{sn_jnl}
\usepackage{subcaption}
\usepackage{graphicx}

\jyear{2021}%

\theoremstyle{thmstyleone}%
\theoremstyle{thmstyletwo}%

\theoremstyle{thmstylethree}%

\begin{document}

\title[Learning user-defined sub-goals using memory editing in reinforcement learning]{Learning user-defined sub-goals using memory editing in reinforcement learning}


\author{\fnm{GyeongTaek} \sur{Lee}}\email{gyeongtaek.lee@rutgers.edu}
\affil{\orgdiv{Department of Industrial and Systems Engineering}, \orgname{Rutgers University, The State University of New Jersey}, \orgaddress{\street{96 Frelinghuysen Road}, \city{Piscataway}, \postcode{08854}, \state{NJ}, \country{USA}}}


\abstract{The aim of reinforcement learning (RL) is to allow the agent to achieve the final goal. Most RL studies have focused on improving the efficiency of learning to achieve the final goal faster. However, the RL model is very difficult to modify an intermediate route in the process of reaching the final goal. That is, the agent cannot be under control to achieve other sub-goals in the existing studies. If the agent can go through the sub-goals on the way to the destination, the RL can be applied and studied in various fields. In this study, I propose a methodology to achieve the user-defined sub-goals as well as the final goal using memory editing. The memory editing is performed to generate various sub-goals and give an additional reward to the agent. In addition, the sub-goals are separately learned from the final goal. I set two simple environments and various scenarios in the test environments. As a result, the agent almost successfully passed the sub-goals as well as the final goal under control. Moreover, the agent was able to be induced to visit the novel state indirectly in the environments. I expect that this methodology can be used in the fields that need to control the agent in a variety of scenarios.}




\maketitle

The goal of reinforcement learning (RL) is to maximize a reward that an agent receives in a specific environment. For example, the RL in robotics allows a robot arm to perform the desired task, such as picking up an object \cite{sinha2022s4rl}. In a game environment such as Atari, the RL allows the agent to get a maximum game score \cite{mnih2013playing}. In sports such as Alphago \cite{silver2017mastering}, the RL is aimed to win the game. In a path planning problem \cite{bae2019multi}, the agent learns to move the shortest distance from the starting point to the target point in the RL. In this study, as well as learning the final goal of the agent, I propose a methodology to learn sub-goals that are defined by the user using memory editing. Let’s assume that we want to reach the destination. The most important thing is to move the shortest distance from the starting point to the target point. However, we usually encounter various situations such as traffic jams or the vagaries of the weather, so we must consider a number of variables that can actually happen. In reality, we can change the intermediate stop or re-plan our route from the scratch depending on the situation. At this point, I focus on making the agent reach the destination by changing the user-defined path by editing the agent's memory. That is, the main purpose of this study is to make the agent being controlled in the process of the agent performing the policy. While most of the RL studies have paid attention to maximizing a reward to attain the desired policy, I propose a methodology that can control the agent of the learned RL model by making the agent learn various sub-goals. As a result, the agent can achieve various sub-goals as well as the final goal in the test environment. \\
In various fields, to learn the agent well in the RL, it is indispensable to give enough rewards to the agent. In addition, it is important for the agent to explore a novel state and exploit previous experience. These are the main challenges for the RL, and various studies have been proposed to address these problems. As for the exploration, many researchers have developed exploration bonus methods \cite{ostrovski2017count,bellemare2016unifying,fox2018dora,machado2018count,silvia2012curiosity,pathak2017curiosity,burda2018exploration}. 
As for the exploitation, several studies have proposed a method to use a replay memory so that the agent stores transitions and uses them efficiently \cite{mnih2013playing, mnih2015human}. Variants of the replay memory were also introduced to improve the exploitation performance \cite{schaul2015prioritized,andrychowicz2017hindsight,nguyen2019hindsight}. Actor-critic algorithms that utilize the replay memory were also proposed \cite{wang2016sample,gruslys2017reactor, oh2018self}. In addition, some studies have recently been introduced to balance the exploitation and the exploration \cite{kim2020take,luger2018dynamic,kang2020balancing, wilson2021balancing}.\\
Meanwhile, several studies on the agent’s multi-goal have been proposed \cite{veeriah2018many, bai2019guided,  lee2020weakly, pitis2020maximum,okudo2021subgoal, kim2021landmark}. These studies generated the multi-goals or sub-goals to enhance the efficiency of the learning for the final goal. The hindsight experience replay (HER) \cite{,andrychowicz2017hindsight,nguyen2019hindsight} improves the exploration performance to utilize a pseudo reward for non-successful trajectories. They used concatenation of the sub-goal and the state similar to the present study to improve the performance of the RL. The existing RL methods for learning the sub-goals have focused on the improvement of the exploration performance and the achievement of one single final goal. However, I focus not only on the achievement of the final goal but also on the achievement of the user-defined sub-goals. The sub-goals are trained simultaneously with the final goal, and the sub-goals are customized by the users on their own. That is, the agent that completed the learning can be controlled by the users. To do this work, I introduce learning the sub-goals using memory editing\\
Memory distortion is common in our daily lives \cite{schacter2011memory}. In essence, all memories are slightly distorted due to various factors such as sleep, retrieval conditions, and conceptual issues \cite{schacter2011memory,bernstein2009tell,fernandez2015benefits}. When we recall the past, we reconstruct them based on what we know and what we have experienced. That is, we often make decisions based on the distorted memory. 
While doing that, it can cause mental illnesses such as trauma, but it can also preserve the information that we perceive and we have experienced \cite{fernandez2015benefits}. If we can artificially edit our memories, we will be able to get away from mental illnesses and concretely learn clearer and essential information \cite{phelps2019memory}. Also, if we can divide and edit the long memory and remind the divided memory intensively, we can get vivid and clear memories. Likewise, if we edit the memory of the agent, we can make the agent perceive the sub-state and learn the various sub-goals.\\
In this study, using the concept of the memory editing, I propose a methodology that the agent learns the sub-goals. Through the memory editing, the agent can learn and recognize the sub-goals and can be controlled by the users in the test environment. In order to do this, I generate various sub-goals in the memory editing and separate learning the original goal and learning the sub-goals by distinguishing two replay memories: one is for learning the original goal and the other is for learning the sub-goals. As the episodes progress, the transitions of the agent are edited by specific probability and the transitions are stored in the second memory. When the experiences of the agent are randomly sampled from the second memory, the agent receives the state of the sub-goals as a state vector, and additional rewards are given with the original rewards. The transitions sampled from the second memory will be different from the learning stages. Because the transitions were edited for learning the sub-goals. It is to make the agent perceive real experiences and reminiscence differently when the agent is trained. This means that when the agent reminds the states and the rewards are given, a sign of the sub-goals is given as a state and the sub-goals are learned with rewards by the sign. The proposed methodology allows the agent to achieve not only the final goal but also the user-defined sub-goals. The agent can perform the learned policy under the user’s control. I applied the proposed methodology to simple environments and confirmed that the agent could reach the final goal, passing the sub-goals which are customized by the user. The main contributions of this article are as follows:
\begin{itemize}
\item The proposed methodology is very simple and easy to implement. We utilize the replay memory widely used for the exploitation and learn the final goal and the sub-goals separately. \\
\item We propose a methodology to indirectly improve the exploration performance using the memory editing. As the agent recognizes the states of the sub-goals in the learning and after that, the states are given to the agent in the episode, the agent is induced to explore the novel states. \\
\item To the best of our knowledge, this study is the first RL methodology in that the agent is controllable such that it achieves the user-defined sub-goals as well as an original goal. By using this methodology, users could define a lot of sub-goals in the real environment in addition to the achievement of the final goal. 
\end{itemize}

\section{Result}\label{sec2}



\subsection{Environment}\label{sec3}
\begin{figure}[H]
\centering
\includegraphics[scale=0.45]{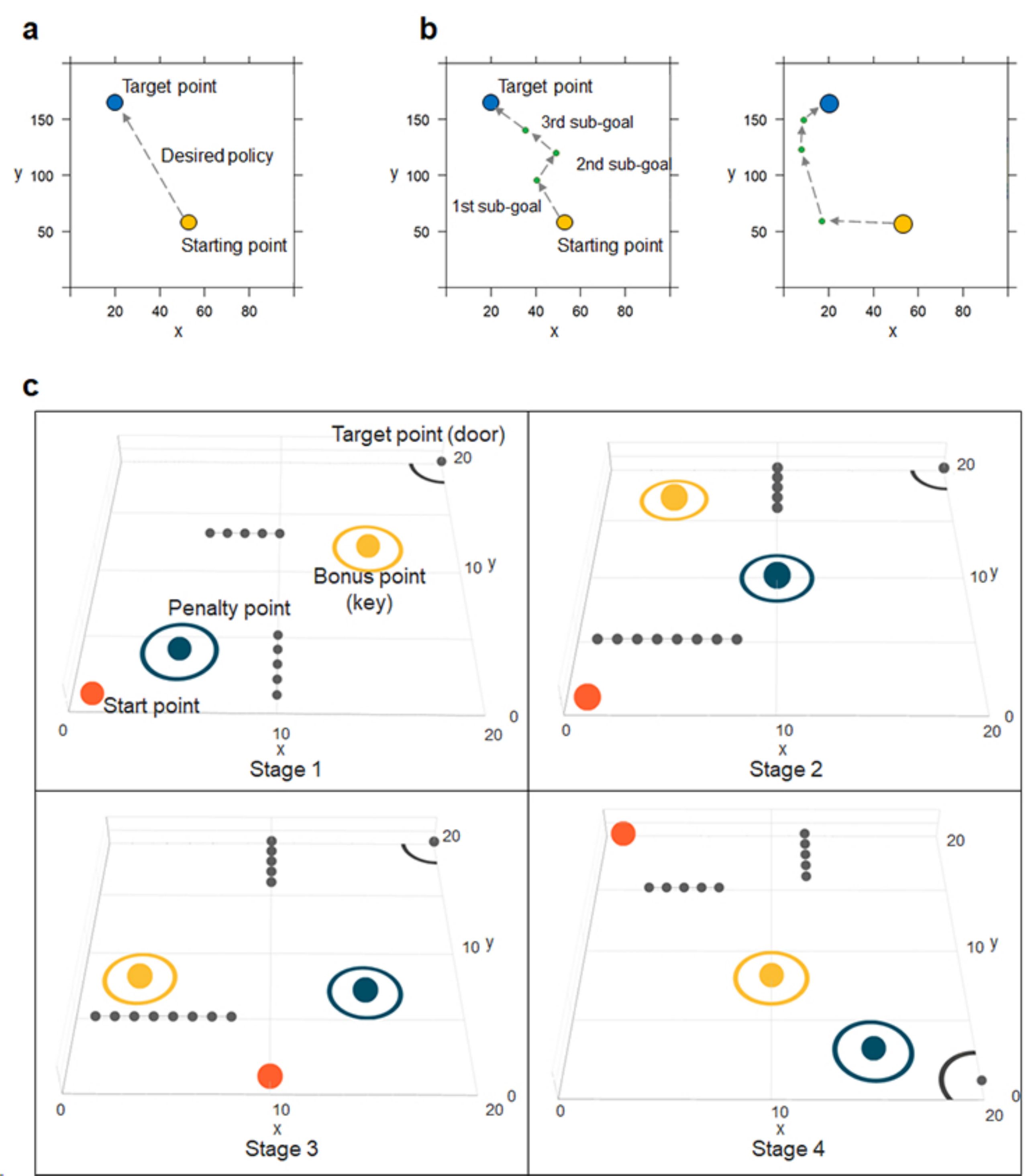}
\caption{Experimental environments. \textbf{a}, Simple 2D grid environment. In the environment, the goal of the agent was set to reach the target point.  \textbf{b}, Examples of scenarios in the 2D grid environment. Three sub-goals were given to the agent, the goal of the agent was to pass the sub-goals and reach the target point. \textbf{c}, Key door domain environment. The environment consisted of a total of 4 stages, and the agent should pass the bonus point to clear each stage.  }
\label{fig2}
\end{figure}

I conducted two simple experiments to see the agent achieve both the sub-goals and the final goal. In these experiments, I assumed various scenarios and wanted to confirm that the agent can be controlled by the user and that the agent behaves differently with and without the existence of the given sub-goals. In the first experiment, I constructed a simple two-dimensional (2D) environment as shown in Figure \ref{fig2}.a. The final goal is for the agent to move from the starting point to the target point. I represented the coordinates of the agent as a state using an effective coordinate vector \cite{lee2020autonomous}. The action of the agent was set as a simple movement: left, right, up, and down. The reward was set as zero except for when the agent reaches a target point or moves out of the grid environment. I constructed the environment as a sparse reward environment to show that the proposed methodology can improve the performance of the exploration. After training, I set up various scenarios. Figure \ref{fig2}.b shows the examples of the test environments. The three sub-goals were imposed on the agent, and the agent should pass the sub-goals and reach the target point. \\
In the second environment, I constructed a ‘key-door domain’ as a hard sparse reward environment as shown in Figure \ref{fig2}.b. The environment consisted of a total of 4 stages. In each stage, although the agent goes to the goal (door), if the agent does not pass the bonus point (key), the agent cannot move to the next stage. To clear the current stage, the agent must pass the bonus point above all. Furthermore, each stage has different positions on the walls, the starting point, the bonus point, the penalty point, and the goal point. Thus, it is very difficult for the agent to clear all stages. The reward was set as zero except for when the agent reaches the bonus point (+10), the penalty point (-10), and the target point (+20). 

\subsection{Experimental settings}\label{sec4}
In the proposed methodology, the memory editing and the exploitation are the most important parts for the agent to learn the sub-goals. Thus, it is inevitable to use the replay memory for the exploitation of the sub-goals. The self-imitation learning (SIL) only exploits valuable past decisions compared to the current value \cite{oh2018self}. The exploitation technique can help the exploitation of learning the sub-goals. Therefore, I utilized the SIL as a base architecture of the RL model. Further, I used the random network distillation (RND), which is widely used as an exploration bonus method \cite{burda2018exploration}. \\
The probability of performing the memory editing and learning the sub-goals is adjusted by using an e-greedy method. I set the probability of how often the memory editing is performed as 0.1. Also, the memory editing is performed when the total reward of the current episode is in the top 1\% of the last 1,000 episodes. In addition, the interval of the sub-goals was set as the number sampled between 5 and 100 for the first experiment and between 5 and 30 for the second experiment. The exploitation of the sub-goals is rarely performed in the earlier stage of the learning. The probability of learning the sub-goals was set as 0.001 in the initial episode. The probability was gradually increased to 0.5 in the final episode. \\
I learned the agent 50,000 episodes for the two experiments and repeated each experiment 10 times. Also, I assigned a coordinate of the location to be reached by the agent as the state of the sub-goals. \

\subsection{Experimental results}\label{sec4}
\begin{figure}[H]
\centering
\includegraphics[scale=0.45]{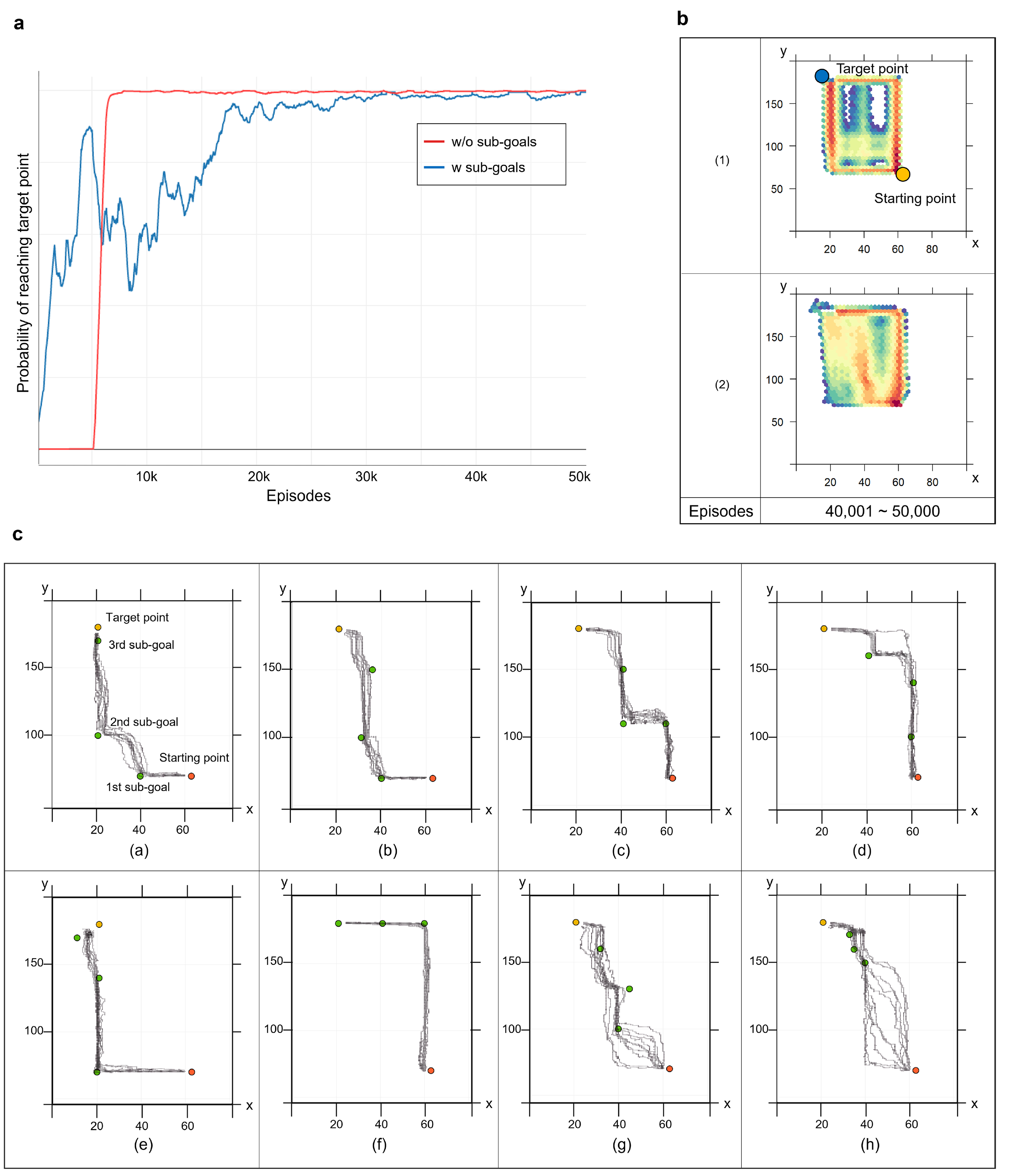}
\caption{The result of learning the sub-goals. \textbf{a}, The probability plot of reaching the target point with and without the learning the sub-goals. \textbf{b}, The visualization of the path of the agent with (2) and without (1) learning the sub-goals. The agent used various paths to reach the target point by learning the sub-goals in the last episode interval (2). \textbf{c}, The visualization of the path of the agent with the given sub-goals. I set the various sub-goals to confirm the performance of the agent. The agent almost successfully passed the sub-goals and reached the target point.}
\label{fig3}
\end{figure}

Figure \ref{fig3}.a shows the plot of the probability of reaching the target point with training the sub-goals and without training the sub-goals in the first experiment. The result shows that the agent reached the target point within 5,000 episodes regardless of training the sub-goals. Figure \ref{fig3}.b is the visualization of the route of the last 10,000 episodes with training the sub-goals (2) and without training the sub-goals (1). The color change from blue to red indicates the frequency of the agent’s visit. It is shown that the agent with training the sub-goals visited wider areas compared to those without training the sub-goals. When the agent started to perceive the sub-goals in the learning and if the agent is given a random sign that makes the agent move to another location, the agent visited the novel state. In addition, without training the sub-goals, the agent moved along the paths that are not varied as the learning progressed, as shown in Figure \ref{fig3}.b.(1). On the other hand, with training the sub-goals, the agent used a variety of routes, as shown in Figure \ref{fig3}.b.(2). We can see that learning the sub-goals indirectly drives the agent to explore the novel state further. \\
Figure \ref{fig3}.c is the visualization of the route of the agent that is given the sign of the sub-goals. The green dot points are the sub-goal points. The agent gets a sub-goal sign nearest to the current location. If the agent reaches a given sub-goal or passes the sub-goal, the agent gets another sub-goal sign nearest to the current location except for the previously assigned sub-goals. I assumed various scenarios and set the sub-goal points between the starting point and the target point differently for each test experiment. In the result, I confirmed that the agent almost successfully moved to the sub-goal points that I customized. It is very interesting to see that after the learning, the agent can pass the user-set middle point and the agent can be controlled by the users using the sub-goals. \\

\begin{figure}[H]
\centering
\includegraphics[scale=0.40]{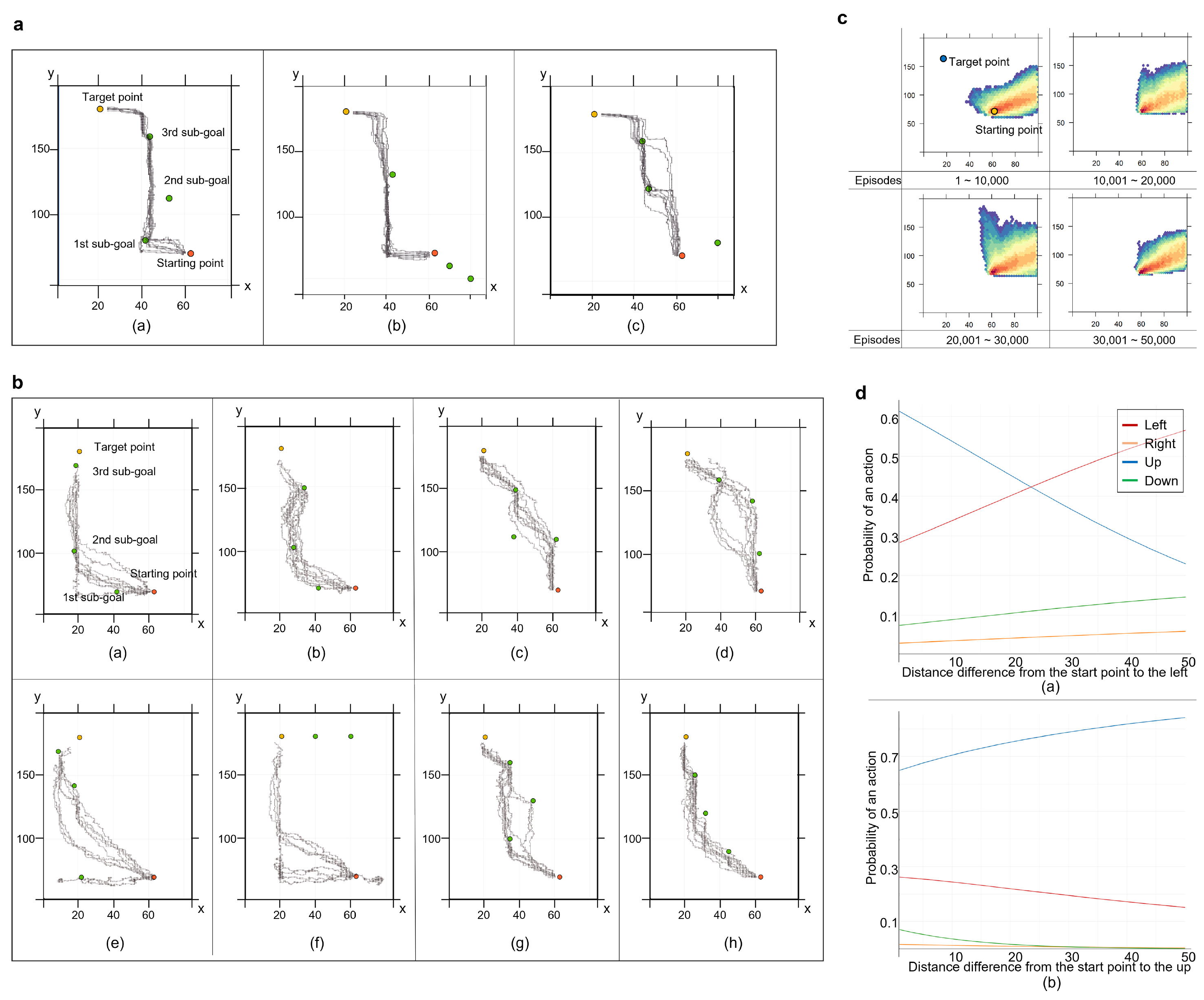}
\caption{\textbf{a}, Failure cases of going through almost all the sub-goals. The sub-goals were set in a difficult way for the agent to pass. However, the agent passed the rest of the sub-goals and reached the target point. \textbf{b}, The visualization of the paths by which the agent went through the sub-goals but failed to learn the original goal. The agent was not able to reach the target point in the learning, the agent reached the target point through the sub-goals. \textbf{c}, The visualization of the path by which the agent failed to learn the original goal. \textbf{d}, The probability plot of the action when the sub-goals were given. }
\label{fig4}
\end{figure}

Figure \ref{fig4}.a show the cases where the agent fails to pass all of the sub-goals. The sub-goals were difficult for the agent to pass. Because the agent’s task was to reach the goal point and the agent was not trained to move the sub-goals in the learning, the agent does not need to zig-zag or go back a long way. However, although the agent could not pass the first sub-goal, the agent tried to move to the next sub-goal and finally, the agent reached the target point. This result means that if the agent cannot achieve the sub-goal, the agent will try to achieve the next sub-goals and the final goal. This study can be helpful for a variety of test environments as we can assume various scenarios using the sub-goals and the final goal. By using various experimental scenarios and the results, we can decide the optimal policy of the agent depending on the situation where there are a number of variables and the agent should be under control. \\
Figure \ref{fig4}.b $\sim$ c shows a case that failed to learn the final goal. In this experiment, the agent never reached the target point (c). However, as shown in Figure \ref{fig4}.b, the agent nearly passed the sub-goals and attempted to reach the target point in the test environment. Of course, the agent could not reach the goal point clearly just like the case in the previous experiment. The agent reached the closest sub-goal and tried the next sub-goal step by step. This is a very impressive result that the agent was able to reach the destination through the sub-goals despite the failure of learning the final goal in the training. The implications of these results are very significant. In this experiment, even though I used a technique for the agent to reach the target point using the sub-goals in the test experiment, I never used it in the training environment. If we can utilize the ability for reaching the agent’s sub-goals, we can easily learn the agent to achieve the final goal like the previous studies \cite{andrychowicz2017hindsight,nguyen2019hindsight,pitis2020maximum,okudo2021subgoal,kim2021landmark}.\\
Figure \ref{fig4}.d shows the probability of the actions at the starting point according to the state of the sub-goals. In the top of figure, the further the sub-goal was set to the left of the starting point, the higher the probability of a 'Left' action was. Likewise, In the bottom of figure, the further the sub-goal was set to the up of the starting point, the higher the probability of an 'Up' action was. These results were usually found in the area visited by the agent. It means that if the agent is given the specific location as suggested in the sign of the sub-goals, the agent will try to reach the sub-goals. That is, by using several sub-goals, the users can control the agent in various scenarios.\\

\begin{figure}[H]
\centering
\includegraphics[scale=0.5]{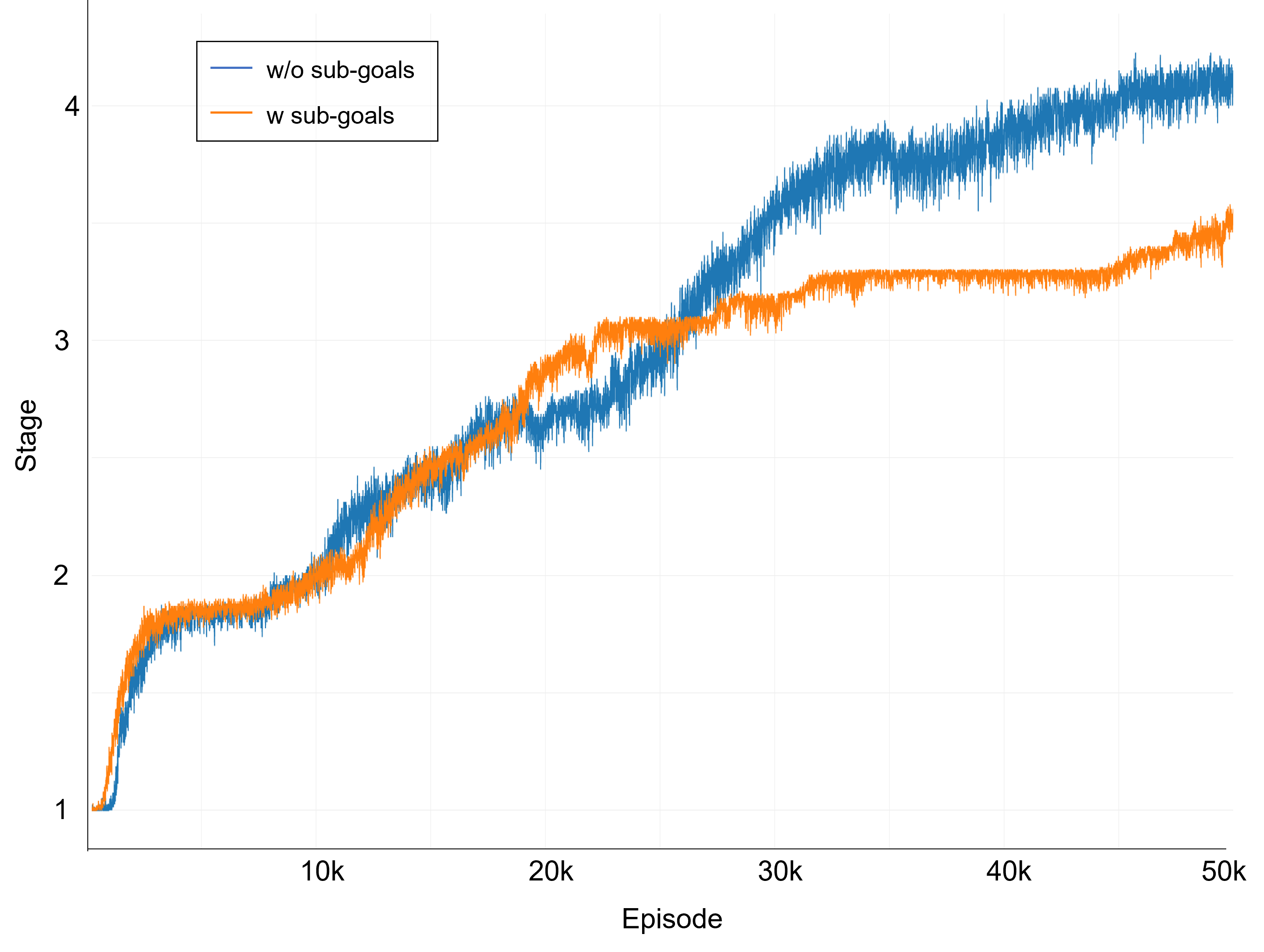}
\caption{Averaged stages for 10 experimental stages with and without learning the sub-goals. In the early stage of the episodes, the original RL showed a better performance. However, as the episode progresses, the agent that learned the sub-goals cleared more stages.    }
\label{fig5}
\end{figure}

\begin{figure}[H]
\centering
\includegraphics[scale=0.3]{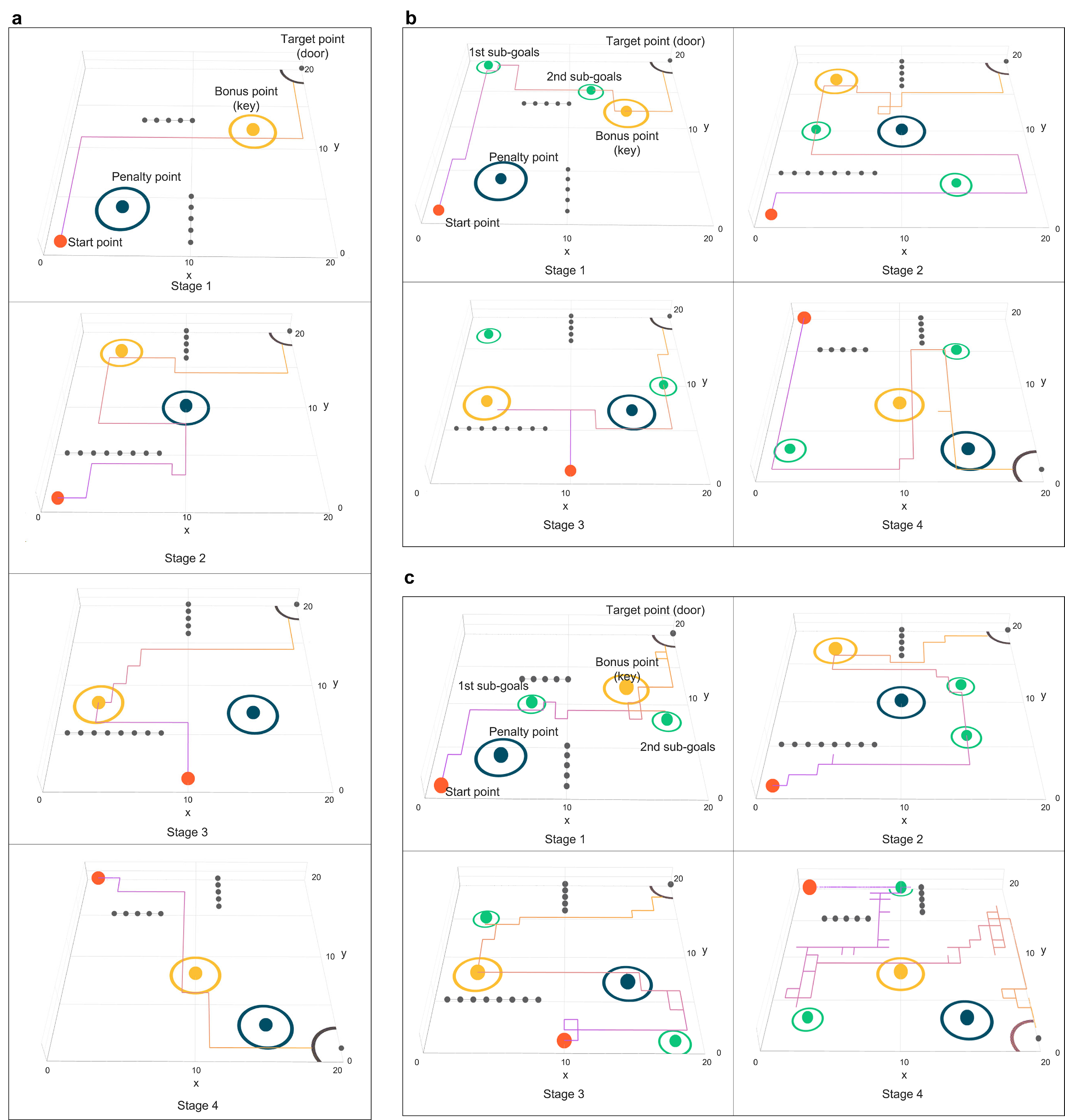}
\caption{\textbf{a}, The visualization of the agent in the second environment without learning the sub-goals. The agent cleared all stages from almost the shortest distance. \textbf{b}, The visualization of the agent with learning the sub-goals in the first scenario in the test environment. \textbf{c}, The visualization of the agent with learning the sub-goals in the second scenario. The agent passed the sub-goals, the bonus point, and the target point for all stages. However, the agent sometimes ignored the sub-goals or passed the penalty point.}
\label{fig6}
\end{figure}

In the second experiment, the algorithms such as prioritized experience replay, Sample efficient actor-critic With experience replay, and SIL without the RND never passed Stage 2 for 10 times in the experiment \cite{schaul2015prioritized, wang2016sample, oh2018self}. In the SIL with the RND (SIL + RND) but without training the sub-goals, the agent passed all stages 1 out of 10 times. Otherwise, in the model with training the sub-goals, the agent passed all stages 5 out of 10 times. I confirmed the result in Figure \ref{fig5}. Until the middle of the learning, the agent without learning the sub-goals reached Stage 4 faster. However, as the episode progresses, the agent with learning the sub-goals started to clear the final stage. This result means that training the sub-goals can affect the exploration of the agent indirectly so that agent could clear the difficult stage. After the learning is completed, it was confirmed that the agent reaches the target point through the bonus point in the almost shortest path for all stages without the sign of the sub-goals, as shown in Figure \ref{fig6}.a.
In the test environment, I assumed the two scenarios as shown in Figure \ref{fig6}.b and Figure \ref{fig6}.c. Two sub-goals were set differently at each stage, the bonus point was set as a third sub-goal, and the sub-goals were granted as the state of the sub-goals to the agent, similar to the previous experiment. As a result, the agent almost passed the sub-goals and attempted to reach the target point. However, in this experiment, the agent could not move the shortest path, whereas the agent that did not learn the sub-goals showed the almost shortest path. In addition, oftentimes, the agent passed the penalty point as shown in Figure \ref{fig6}.c.(Stage3). I observed that the phenomenon occurred sometimes when the agent learned the sub-goals. This environment is a relatively small world compared to the first experiment. Thus, it is hard for the agent to transform the direction within the short term. Furthermore, the agent trained to pass the bonus point to clear each stage, but if the sub-goals were given to the agent, the agent was likely to be confused about whether to go the sub-goal, bonus point, or target point within a short period. Here, future research is needed for the agent to enforce the given sub-goals clearly.

\section{Discussion}\label{sec4}
In this article, I proposed a methodology to train the sub-goals using the memory editing so that the agent can reach the sub-goals as well as the final goal under control. The results of the two experiments were very impressive, which can be useful in real test environments. 
To learn the sub-goals, I used the memory editing and the separated exploitation method. I distinguished learning the sub-goals from learning the final goal. The transitions collected from the episode are transformed into the sub-goals in the memory editing, and the intervals between the sub-goals and the additional rewards are given. At the beginning of the learning, the final goal is learned at first. As the learning progresses, learning the sub-goals is performed gradually. Learning the sub-goals is separated from learning the final goal using different replay memory and the value network. As a result, the agent started not only to learn the final goal but also slowly started to learn the sub-goals. I conducted experiments on the various scenarios for the sub-goals. The experimental results show that the agent could successfully reach various sub-goals that are customized by the user as well as one final goal. Even though the agent failed to reach the final goal in the learning, in the test environment, the agent could arrive at the final destination through the sub-goals. This result is very significant and can be contributed to various domains and studies. Especially, this methodology will be helpful in the fields that need to control the agent in various scenarios or situations. \\ 
However, in this study, there are several limitations to solve. First, it needs to have enough episodes to learn the sub-goals. Because learning the sub-goals should be performed after the final policy is learned enough. Second, it is hard to adjust a balance between learning the final goal and the sub-goals for controlling the agent completely. If the exploitation of the sub-goals is performed too often, the agent can be fallen into the local policy. Meanwhile, if the exploitation of the sub-goals is rarely performed, the agent cannot recognize the state of the sub-goals. Finally, the agent can reach almost only the sub-goals that had been previously visited by the agent. This problem is possibly due to a natural phenomenon, which still needs to be solved for the agent to be fully controlled. Learning the sub-goals is performed by the exploitation of the transitions collected from the previous episodes. Therefore, the agent was able to move almost only the sub-goals visited by the agent. However, in the real environment, unexpected situations can occur at any time. Future research is needed for the robust RL model to learn the unexpected sub-goals. I expect that this methodology will be developed in the future and applied to various problems and domains that need to control the agent. \\

\section{Method}\label{sec4}

This study was motivated by the following question: How can we learn the agent to perform various tasks under control as well as to achieve the original goal? In general, the agent is learned to achieve the user-defined goal by maximizing the total sum of rewards or to maximize the rewards given from a specific environment such as a game. In the test environment, the agent that has completed learning merely performs to get the rewards based on the policy network. When we drive a car to reach the goal, we should consider a number of situations such as ‘is there a pedestrian on the street?’, ‘where is a crosswalk?’, and ‘what direction does that sign indicate?’. In a complex driving environment, we generally get some directions for the sub-goals and are aided by many operations such as 'deceleration', 'stop', and 'rotation' via a navigator. Here, we can select and perform many sub-goals on our own to achieve the goal. Our motivation is that humans can manipulate and perform the sub-goals for the achievement of the final goal. Therefore, I propose a methodology to perform the user-defined sub-goals by using the concept of the memory editing. \\

\subsection{The memory editing of the agent}\label{sub_memorye}
The HER \cite{andrychowicz2017hindsight} used the sub-goals for the agent to achieve the goal faster. However, the purpose of the sub-goals and the training process of the HER are different from the present study. The HER focused on training agents more efficiently and faster towards their goals. They used the sub-goals and the states together by concatenating the elements for all the training procedures. Meanwhile, I focus on training the sub-goals to allow the agent to perform the sub-goals and the final goal under control using the memory editing. The sub-goals can be customized by the users. In addition, I distinguish training the sub-goals from the training of the final goal by using two replay memories. Because it is possible that the transitions for the sub-goals can be confused with the transitions for the original goal after performing the memory editing. Also, it is difficult to adjust to what extent each learning can be performed. If the sub-goal is not trained properly, the agent is more likely to fall into the local policy. \\
To train the sub-goals, training the final goal should be preceded to some extent. As the episode progress, training the sub-goals should be gradually performed. When one episode ends, the states in the episode are transformed into the state of the sub-goals by a specific probability and stored in the second memory. This process is called memory editing. At this point, it is very important how often and under what conditions to edit the memory. As mentioned previously, if the sub-goals are trained often, the agent can be fallen into the local policy. Thus, it is reasonable to increase the frequency of the memory editing slowly as the episode progress. Moreover, we need to select valuable sub-goals in the memory editing. It is preferable to choose the sub-goals within the subset of the processes from the initial state to the goal, but it does not necessarily mean that it is a sufficient condition. I propose to generate the sub-goals when the total sum of the reward of the episode is higher than previous episodes or to generate randomly with a low probability. \\

\subsection{The exploitation of the sub-goals}\label{sub_memory}
\begin{figure}[H]
\centering
\includegraphics[scale=0.37]{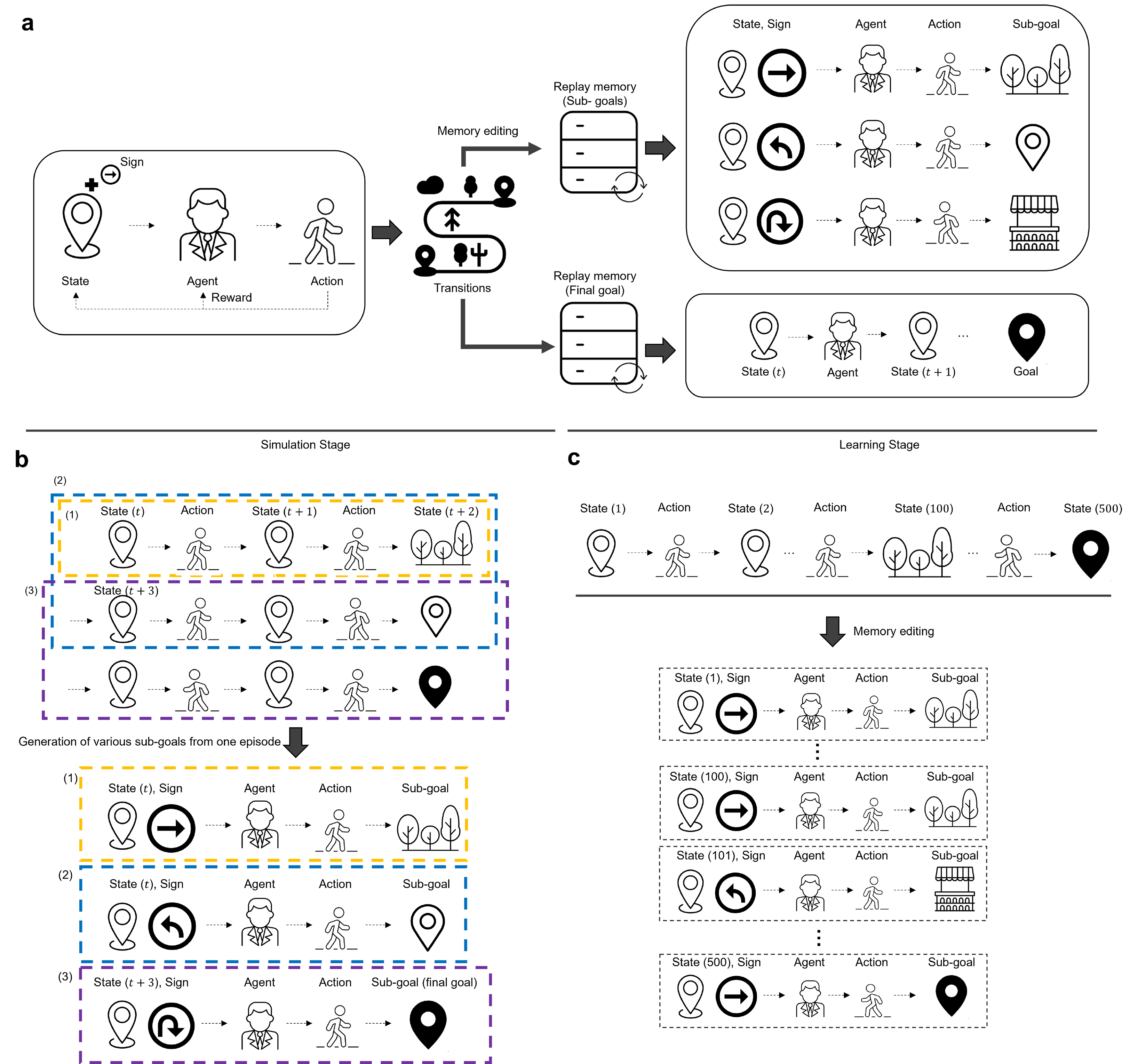}
\caption{\textbf{a}, An overview of the proposed methodology. In the simulation stage, the transitions are collected, and when one episode is completed, the memory editing has been performed. In the learning stage, learning the sub-goals and learning the original goal are progressed separately. \textbf{b}, An example of the memory editing. In one episode, depending on the interval of the sub-goals, various sub-goals can be generated. \textbf{c}, An example of generating the sub-goals. When the interval of the sub-goals is 100, the 100th state is given to the previous 99 states.}
\label{fig1}
\end{figure}

After the episode stage, the states of the sub-goals and the additional rewards are given to the agent in the memory editing. Then, the transitions should be stored in the second memory and should be trained using the replay memory. However, there are two problems to solve. First, the state of the sub-goal has a different characteristic from the original state. The value network is used to evaluate the original states. As a result of the evaluation, the probability for the transitions to be sampled is calculated. The reason is that the state of the sub-goal is fundamentally different from the original state, it is not reasonable to use the value network. Thus, I utilize an additional value network to evaluate the value for the states of the sub-goals. The value network for the sub-goals is only used to evaluate and decide whether the state of the sub-goals is being exploited or not. In this study, this network will be referred to as a value network for the sub-goals ($VN_s$), and the original value network, as a $VN$. Second, as previously mentioned, using one replay memory can cause a decrease in the efficiency of learning. The first aim of the RL is to learn the agent to achieve the final goal. Also, the capacity of the memory is limited, and the states of the sub-goals need to be reevaluated by the $VN_s$. Therefore, after the memory editing, the transitions transformed for the sub-goals should be stored separately in the replay memory and trained gradually. Thus, I employ another replay memory to store the edited memories and learn the sub-goals. It is important for this study to manage and supervise the edited memories individually for the sub-goals. The memory will be referred to as a replay memory for the sub-goals ($RM_s$), and the original replay memory for the final goal, as an $RM$. By utilizing the $RM_s$ for the exploitation, learning can become efficient.\\
Figure \ref{fig1}.a and Algorithm 1 shows the process of the proposed methodology. In the episode stage, almost all the simulation process is similar to the existing RL method. The agent performs based on the current policy network ($\theta_{p}$) and gets the next state ($s_{t}$) and the reward ($r_t$) until the end of the episode. One of the crucial points different from the original RL framework is to insert a sign for the agent to explore the novel state. The agent is learned for the final goal at first, and is gradually learned for the sub-goals. As the agent starts to recognize the sub-goals in the training, if the agent gets the sub-goals $s_{t} \| g$ as a state just like a sign in the traffic, the agent can explore a state that has not been visited in the previous episodes. The sub-goal is given to the agent with small probability.\\
When one episode finishes, whether the memory editing is performed by a specific probability is decided in the transition. Then, if the sub-goals are generated in the memory editing, it is necessary to decide which sub-goal to be chosen in the episode. The sub-goals can be closer goals for the short term and distant goals for the long term. If we want the agent to achieve a distant goal, the interval of the sub-goals to be generated in the steps of the episode need to be divided into several chunks. Figure \ref{fig1}.b shows various sub-goals being generated in one episode. Depending on the interval, the sub-goal of the specific point in the episode can be a close branch (1) or the goal (3). Figure \ref{fig1}.c shows the process of generating the sub-goals in the memory editing. Let’s assume that the size of steps in the episode is 500 and we want to make the agent achieve the sub-goals for the long term. Then, we can generate the sub-goals every 100 steps in the episode. And the 100th state is given to the previous 99 states as a state of the sub-goal as shown in Figure \ref{fig1}.c. The sub-goals are represented as the state and concatenated with the original state $s_{t} \| g_{int}$ . If we want to make the agent reach the sub-goals near the current step, we can impose the sub-goals with short intervals like 5, 10, or 20.\\
In the learning stage, the transitions are trained using the $RM$ to achieve the final goal. Then, as the episode progresses, the sub-goals are sampled by the probabilities that are calculated by the $VN_s$ and are trained using the $RM_s$ gradually. It is difficult to decide how often the sub-goal is trained. 	At least, the sub-goals should be learned after the final goal is learned enough. If the sub-goals are learned frequently in the earlier episode, the agent can be fallen into the local policy. Therefore, the exploitation of the sub-goals should be performed when the final goal is almost achieved. 

\begin{algorithm}[H]
\begin{algorithmic}[1]
\caption{Learning the sub-goals using memory editing }\label{alg:algorithm1}

\State Initialize policy network parameters $\theta_{p}$
\State Initialize replay buffer for original goal $\mathcal{RM} \leftarrow \emptyset$
\State Initialize replay buffer for sub-goals $\mathcal{RM_s} \leftarrow \emptyset$
\Procedure{Learning the Sub-goals and the final goal}{}
\For{episode = 1, M}
\State  \verb|\\| Simulation stage.
\For{each step}
\State Generate a random sub-goal $s_{t} \leftarrow s_{t} \| g$  with small probability
\State \algorithmiccomment{$\|$ denotes concatenation}
\State Execute an action $s_t,a_t,e_t,s_{t+1} \approx \pi_{\theta}(a_t \mid s_t)$
\State Store transition $\mathcal{E}\leftarrow \mathcal{E} \cup \{(s_t,a_t,r_t)\}$
\EndFor

\If{ $s_{t+1}$ is terminal}
\State Compute returns $R_t= \Sigma^\infty_{k}\gamma^{k-t}{r}_k$ in $\mathcal{E}$
\State $\mathcal{RM}\leftarrow \mathcal{RM}\cup \{(s_t,a_t,R_t)\}$
\State Clear episode buffer $\mathcal{E} \leftarrow \emptyset$

\EndIf
\State
\If{ $R_t$ $>$ memory editing threshold}
\State Sample interval $int$ of steps in the episode
\State Generate sub-goals $g$ and state of sub-goals $s \| g$
\State Set additional rewards $r_t'$ 
\State $\mathcal{RM_s}\leftarrow \mathcal{RM_s}\cup \{s_t \| g_{int},a_t,r_t')\}$
\EndIf

\State
\State   \verb|\\|Learning stage.
\For{k= 1, N}
\State  Sample a minibatch $\{(s,a,R)\}$ from $\mathcal{RM}$
\State \algorithmiccomment{Optimize policy network  $\theta_{p}$ for the final goal}
\EndFor
\For{k= 1, P}
\State  Sample a minibatch $\{(s \| g,a,r')\}$ from $\mathcal{RM_s}$
\State \algorithmiccomment{Optimize policy network  $\theta_{p}$ for the sub-goals}

\EndFor
\EndFor
\EndProcedure

\end{algorithmic}
\label{algo1}
\end{algorithm}




\bibliography{mybibfile.bib}

\end{document}